\documentclass{optica-article}

\journal{opticajournal} 

\articletype{Research Article}

\usepackage{orcidlink}
\usepackage{graphicx}
\usepackage{soul}
\usepackage{caption}
\usepackage{subcaption}
\usepackage{algorithm}
\usepackage{algpseudocode}
\usepackage{siunitx}

\begin{document}

\title{Single Pixel Image Classification using \\an Ultrafast Digital Light Projector}

\author{Aisha Kanwal,\authormark{1}\,\orcidlink{0009-0003-4257-8847}
Graeme E. Johnstone,\authormark{1}\,\orcidlink{0000-0001-5471-4664} 
Fahimeh Dehkhoda,\authormark{2}\,\orcidlink{0000-0002-7399-7169}
Johannes H. Herrnsdorf,\authormark{1}\,\orcidlink{0000-0002-3856-5782}
Robert K. Henderson,\authormark{2}\,\orcidlink{0000-0002-0398-7520}
Martin D. Dawson,\authormark{1}\,\orcidlink{0000-0002-6639-2989} 
Xavier Porte,\authormark{1,*}\,\orcidlink{0000-0002-9869-7170}
and Michael J. Strain\authormark{1},\orcidlink{0000-0002-9752-3144}}

\address{\authormark{1}Institute of Photonics, University of Strathclyde, 99 George Street, Glasgow, G1 1RD, UK\\
\authormark{2}School of Engineering, University of Edinburgh, Sanderson Building, Robert Stevenson Road, Edinburgh, EH9 3FB, UK}

\email{\authormark{*}javier.porte-parera@strath.ac.uk} 


\begin{abstract*} 
Pattern recognition and image classification are essential tasks in machine vision.  
Autonomous vehicles, for example, require being able to collect the complex information contained in a changing environment and classify it in real time. 
Here, we experimentally demonstrate image classification at multi-kHz frame rates combining the technique of single pixel imaging (SPI) with a low complexity machine learning model. 
The use of a microLED-on-CMOS digital light projector for SPI enables ultrafast pattern generation for sub-ms image encoding. 
We investigate the classification accuracy of our experimental system against the broadly accepted benchmarking task of the MNIST digits classification. 
We compare the classification performance of two machine learning models: An extreme learning machine (ELM) and a backpropagation trained deep neural network. 
The complexity of both models is kept low so the overhead added to the inference time is comparable to the image generation time. 
Crucially, our single pixel image classification approach is based on a spatiotemporal transformation of the information, entirely bypassing the need for image reconstruction. 
By exploring the performance of our SPI based ELM as binary classifier we demonstrate its potential for efficient anomaly detection in ultrafast imaging scenarios. 

\end{abstract*}

\section{Introduction}
Machine vision is a mature technology embedded in autonomous agents, playing a crucial role in industries as distinct as self-driving vehicles and pharmaceutical packaging.
As the demand for fast image classification solutions grows, the operation bandwidth of conventional digital cameras becomes a challenge.
In this context, event based cameras are quickly gaining traction due to the lower amount of information they transmit in dynamically changing scenarios. 
However, their application remains restricted to the same range of wavelengths that their visible to near-IR cameras are sensitive to. 
In contrast, single pixel imaging (SPI) offers a promising alternative to camera-based machine vision both in high-speed applications\cite{Jiao2018, Pawar2019, Duarte2008} and at unconventional wavelengths outside the range of silicon-based CMOS chips\cite{Stantchev2017, Li2025}. 
In the context of machine vision, single pixel image classification (SPIC) offers the advantage of dramatically reducing the hardware complexity on the detection side and an enhanced operation bandwidth achieved via compressed sensing (CS). 

SPI is a technique that captures images using only a single point detector and a sequence of illumination patterns projected onto the object to be imaged. 
Instead of using a conventional camera chip with many pixels, SPI relies on patterned illumination and computational processing to form an image. 
The structured light patterns play a significant role in determining the efficiency and fidelity of the SPI system\cite{Gibson2020}.  
Hadamard patterns are a widely adopted illumination concept\cite{Gibson2020, Sun2023, Johnstone2024, Li2025} because they are orthogonal and binary, and therefore compatible with two-state displays like digital micromirror devices (DMDs). 
In terms of bandwidth, binary SPI systems are limited by how fast the illumination can switch between patterns as the bandwidth of the single detector usually exceeds the one of the remaining system. 
DMDs can generate arbitrary binary patterns at refresh rates of $\sim 10^{4}$ frame per second (fps), limited by the mechanical nature of mirror switching\cite{Johnstone2024, Edgar2019}. 
In SPI, the number of patterns required to reconstruct an image is in the order of $10^{2}$ symbols, which (at DMD switching speed) results in an image generation rate similar to that of common CMOS cameras ($\lesssim 10^{2}$ Hz). 
CS was first proposed in SPI to address this bandwidth limitation\cite{Cands2007, Duarte2008, Katz2009, Stantchev2017}. 
CS is based on algorithmic strategies to reconstruct images from fewer measurements, enabling faster operation without excessively sacrificing image quality. 
Alternatively one can increase SPI bandwidth using faster hardware to generate the patterns, notably microLED arrays have been recently proposed for pattern generation $\sim10^2$ times faster than DMDs\cite{Xu2018, Johnstone2024}. 
Recent developments used microLED arrays for image classification in analog computing schemes\cite{Kalinin2025, Muller2025}, demonstrating the central role that this technology will play in the next generation of optical computing.  


In this work, we focus on direct spatiotemporal classification without reconstructing the original images. 
Several recent works have shown the power of reconstruction-free SPIC. 
In \cite{Cao2021}, raw single pixel measurements acquired from random patterns were directly fed into a deep neural network for post-processing, achieving over 94\% accuracy at less than 2\% of the Nyquist sampling rate. 
In \cite{zhu2020}, a photon-limited SPI system using sparse binary patterns achieved over 90\% accuracy on MNIST with extremely low photon counts using a simple k-nearest neighbor (kNN) classifier.
Finally, in \cite{Manko2025}, a simulation-based SPI system using Hadamard patterns showed that high classification accuracy 96\% on MNIST dataset can be achieved by using only a small fraction of measurements, with a single hidden layer neural network. 
Here, we experimentally implement a SPIC setup for high speed image capture of characters from the MNIST dataset and demonstrate image classification using both an extreme learning machine model and a deep neural network. 
By using high-speed optical projection hardware in the form of a microLED-on-CMOS array, we illuminate each MNIST digit with a sequence of Hadamard patterns and collect the light superposition of each pattern using a fast photomultiplier single pixel detector.  
The generated photo-signal is recorded as time series with a real time oscilloscope and directly classified using a low complexity machine learning model. 
Our experiment achieves accuracies above 90\% at 1.2 kfps, demonstrating high classification accuracy in a free-space optical setup at unprecedented speeds. 
Moreover, we achieve binary (one-vs-all) classification accuracies $>99\%$ using a minimalist extreme learning machine (ELM) model. 
Such task is analogous to anomaly detection in fast changing scenarios. 

\section{Single Pixel Imaging Experiment}
\begin{figure}[htbp]
    \centering    \includegraphics[width=12cm]{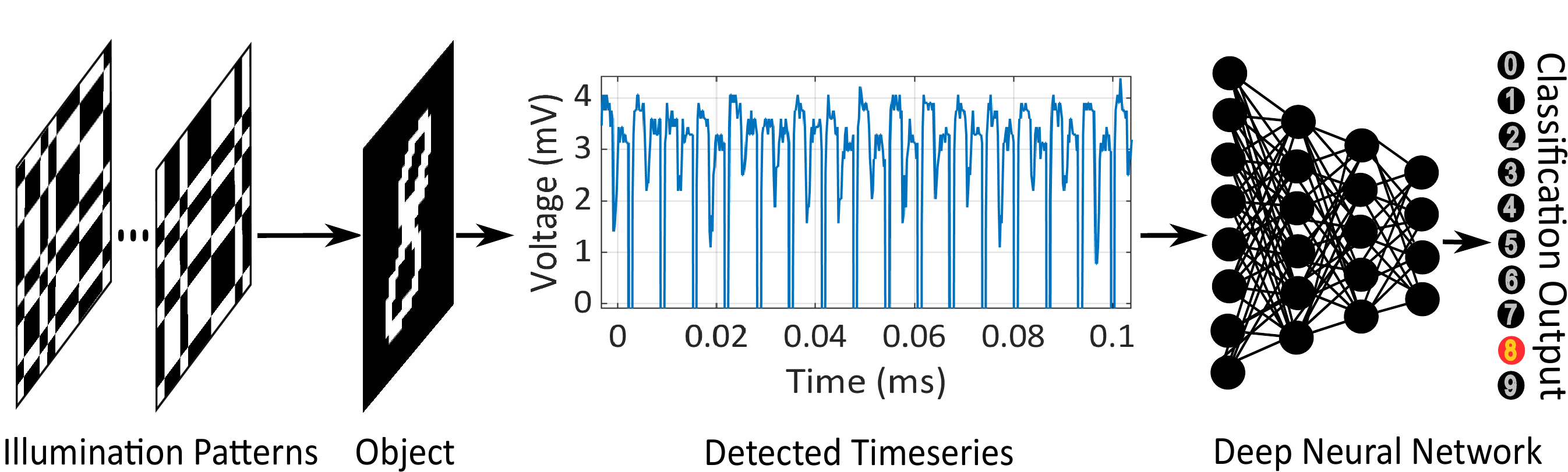}
    \caption{Schematic diagram of our single pixel image classification (SPIC) framework. 
    Each Hadamard pattern in the imaging base is projected onto the object. 
    The superposition of both images is converted into a scalar photocurrent and recorded by a real time oscilloscope. 
    The sequence of all patterns is processed by an artificial neural network whose output is already the object's class.}
    \label{fig:schematicdiagram}
\end{figure}

Figure \ref{fig:schematicdiagram} illustrates our SPIC framework. 
The object to be classified is illuminated with a sequence of binary patterns. 
The time series of photodetected intensities is the result of the superposition of each illumination pattern with the object. 
A low complexity machine learning model is used to map the photodetected intensities onto the task classes. 

We use a Hadamard base to create the structured illumination patterns. 
The Hadamard set forms a complete orthogonal basis of binary patterns derived from the original Hadamard matrices, which are square matrices consisting of elements valued +1 and -1 \cite{Assmus1992}. 
Since negative values cannot be directly represented on a LED-based digital light projector, each Hadamard pattern is split into two separate binary patterns: one for the positive elements and a second inverted pattern for the negative elements. 
These two patterns are displayed sequentially, and the difference in their detected intensities is taken as the measurement. 
The reconstructed image \(I\) can be calculated using the following equation \cite{Gibson2020, Lu2020, Karmakar2023}: 

\begin{equation}
I_{(x,y),M} = \frac{1}{M} \sum_{m=1}^{M} S_m P_{(x,y),m}
\label{eq:reconstruction_image}
\end{equation}
where \(P_{(x,y),m}\) is the m$_{th}$ Hadamard pattern, \(S_m\) is the difference in detected signal between the two reciprocal patterns composing the $m_{th}$ Hadamard pattern, with \(M\) the total number of Hadamard patterns sampled. 
We refer to \ref{appendix:A} for a description of the algorithm used to generate the Hadamard patterns. 

\begin{figure}[htbp]
\centering\includegraphics[width=12cm]{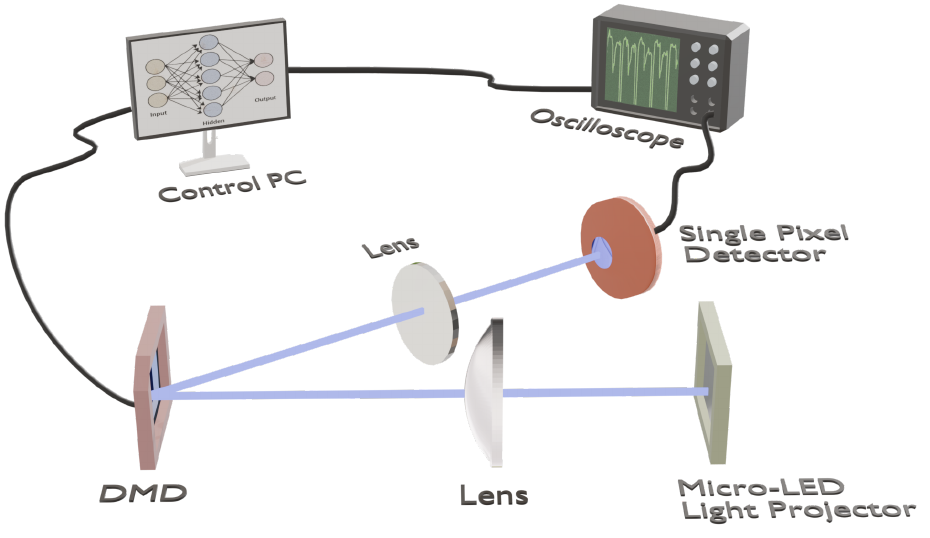}
\caption{Experimental setup for single pixel image classification. 
The microLED projector generates the sequence of Hadamard patterns that illuminate the image presented on the Digital Micromirror Device (DMD). 
The Single Pixel detector collects the superposition of each image pair and transforms it into a scalar photocurrent registered by the real time oscilloscope. 
The full time series corresponding to each image is post-processed by a software encoded ML model.}
\label{fig:experimental-setup}
\end{figure}

The optical setup consists of a high-speed microLED digital light projector, a DMD, a single pixel detector, and an oscilloscope for signal acquisition, as shown in Fig. \ref{fig:experimental-setup}. 
The core of this system is a microLED-on-CMOS digital light projector, which illuminates Hadamard patterns onto the DMD. 
The DMD displays the image to be classified. 
A photodetector "Onsemi Silicon Photomultiplier C-Series 10035", with an active area of $1 \times 1 \mathrm{mm}^2$, captures the intensity of light reflected from the DMD corresponding to each projected pair of patterns. 
This optical signal is then recorded as a time series by a real-time oscilloscope with 1GHz bandwidth.

\begin{figure}[htbp]
\centering\includegraphics[width=13cm]{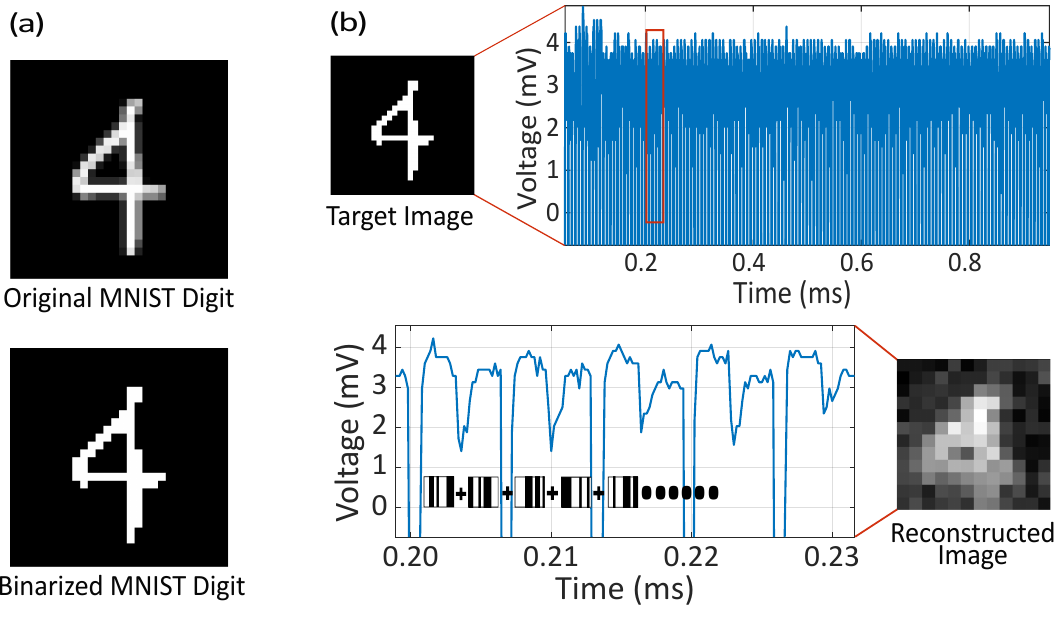}
\caption{Single pixel measurement and reconstruction process. 
(a) Original and binarized MNIST digit used as the target object for structured illumination. 
(b) Illustration of the SPI process: The upper panels show an optically encoded binarized MNIST digit and its corresponding photodetected time series. 
The panels below show a magnification of this time series, superposing the corresponding Hadamard patterns and the reconstructed MNIST digit.}
\label{fig:reconstruction-process}
\end{figure}

The projector features a \(128 \times 128\) active-matrix array of $30 \times 30 \mathrm{\upmu m}^2$ microLED pixels on a 50 $\upmu\mathrm{m}$ pitch, integrated with a smart pixel CMOS driver. 
The projector supports both binary pattern projection and 5-bit grayscale resolution, toggling between stored frames at MHz speeds\cite{Bani2022}.
In our optical system, the microLED projects a structured \(12 \times 12\) Hadamard pattern set (Had12) onto a DMD at a frame rate of 330,000 fps in global shutter mode. 
The size of the Hadamard base chosen here and therefore the total number of patterns is limited by the memory depth of the FPGA board driving the microLED array.  
The DMD used in this setup (Texas instruments DLPLCR70EVM and DLPLCRC410EVM) has a resolution of \(1024 \times 768\) (XGA) with a micromirror pitch of of 13.6 $\upmu\mathrm{m}$ and a diagonal size of 0.7 inches. 
The DMD can change between binary patterns at a nominal maximum rate of 32,552 Hz, displaying a binarized version of the MNIST dataset.  
Figure \ref{fig:reconstruction-process}(a) shows the difference between the original and the binarized datasets. 
The signal acquisition and reconstruction process as enabled by this configuration is illustrated in further detail in Fig. \ref{fig:reconstruction-process}(b). 
The binarized digit "4" is the target digit to be reconstructed, the $\sim 1$ ms time-trace shows the optical signal acquired over time as the full Hadamard patterns set is projected onto the digit displayed on the DMD. 
A zoomed-in section below highlights the type of signal that is recovered from our SPI setup. 
Each plateau corresponds to the superposition of one Hadamard pattern and the object, while the dips between plateaus show the short periods where the microLED display switches off in order to change patterns.   
The inset on the right reveals the reconstructed image, illustrating the validity of our setup for SPI. 

\section{Machine learning models} 
After capturing the optical signals using a real time oscilloscope, our goal is to classify the MNIST handwritten digits based on the time series without reconstructing each image. 
We explore image classification in two different post-processing scenarios: A model inspired in the architecture of extreme learning machines (ELM) and a feed-forward deep neural network (DNN) model trained with error back-propagation.

\begin{figure}[htbp]
\centering\includegraphics[width=13cm]{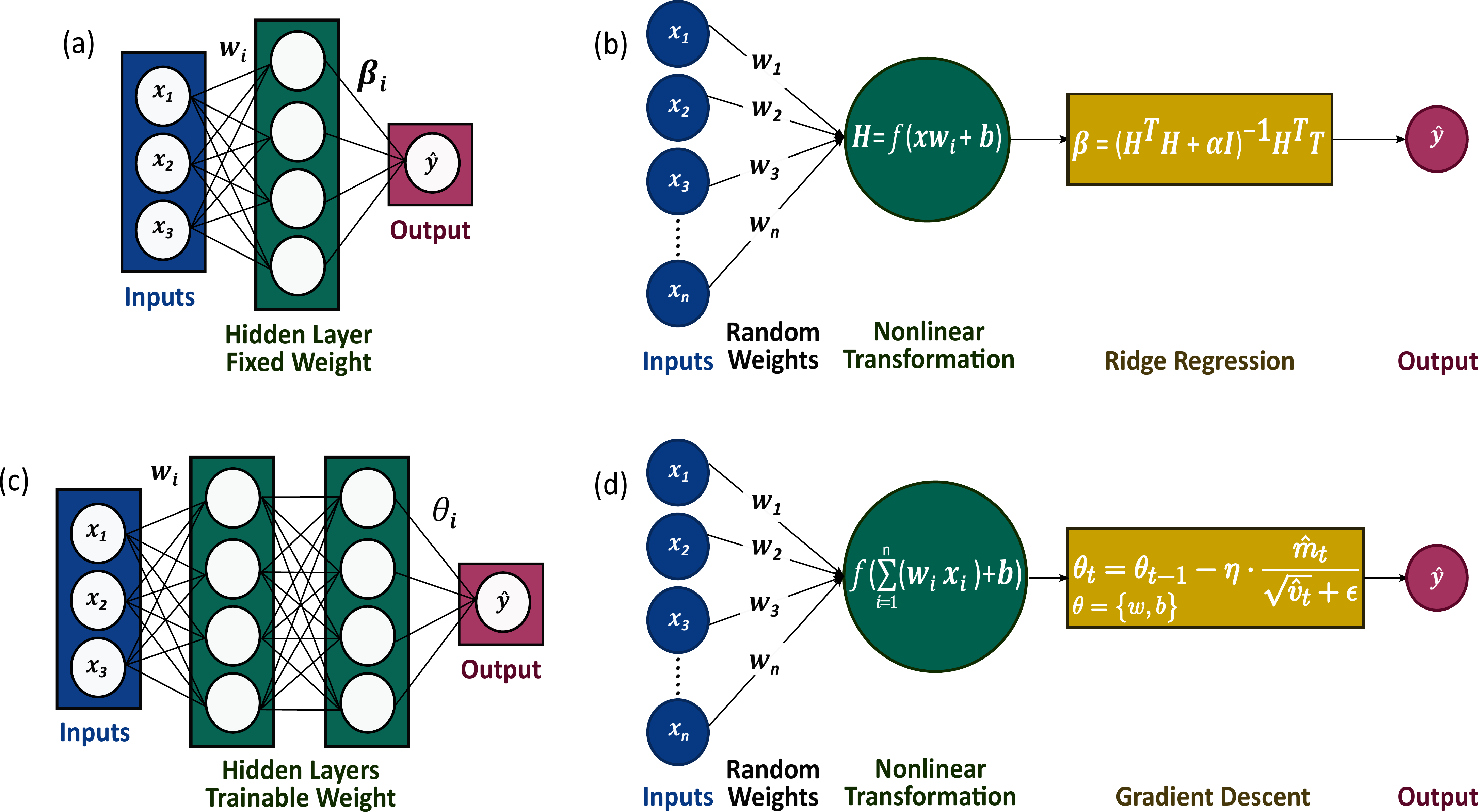}
\caption{Basic architectures and training of the two machine learning models considered in this work. Diagrams (a) and (b) depict the extreme learning machine model used for multiclass and one-vs-all binary classification with fixed random input weights and output training only using ridge regression. 
Diagrams (c) and (d) depict a feed-forward deep neural network for multiclass classification with trainable weights using gradient descent.}
\label{fig:ML-models}
\end{figure}

In the ELM model, we use a single hidden layer neural network in which the input weights are randomly initialized and kept fixed, as depicted in Fig. \ref{fig:ML-models}(a). 
The training strategy is described in the yellow box in Fig. \ref{fig:ML-models}(b). 
Similar to reservoir computing, only the output weights are trained using ridge regression.
Prior to training, the input data \( X \in \mathbb{R}^{N \times d} \), where \( N \) is the number of training samples and \( d \) is the number of input features, is normalized and scaled within the range of \([-3, +3]\). 
The standardized data is then projected to a higher-dimensional space using a randomly initialized weight matrix \( W_{\text{in}} \in \mathbb{R}^{d \times L} \), where \(L\) denotes the number of hidden neurons.
A bias vector \(b \in \mathbb{R}^L\) is added, and the result is passed through a Rectified Linear Unit (ReLU) nonlinear activation function \(\{f(x) = max(0,x)\}\). 
The resulting hidden layer output matrix \( H \in \mathbb{R}^{N \times L} \) is computed as:
\begin{equation}
    H = f(X W_{\text{in}} + b)
    \label{eq:computed_inputs_with_weights}
\end{equation}
To obtain the output weights \( \beta \in \mathbb{R}^{L \times m} \), where \( m \) is the number of output classes, we use ridge regression to minimize the squared error with an \( \ell_2 \) regularization term. 
For \( T \in \mathbb{R}^{N \times m} \) as the one-hot encoded label matrix, the output weights $\beta$ are obtained in one shot using the following regularized least squares solution:
\begin{equation}
\beta = (H^\top H + \alpha I)^{-1} H^\top T,
\label{eq:calculated_output_weights}
\end{equation}
where \(\alpha\) is the regularization parameter, and \( I \) is the identity matrix of size \( L \times L \).
During inference, class score is obtained as: 
\begin{equation}
    Y = H\beta.
    \label{eq:prediction}
\end{equation}
For multiclass classification, the predicted class corresponds to the maximum score: 
\begin{equation}
    \hat{y} = \max(Y).
    \label{eq:multiclass-predicted}
\end{equation}
Here, we also study the binary classification scenario, which is analogue to the problem of anomaly detection. 
For one-vs-all binary classification, a threshold of 0.5 is applied for class prediction. 
The predicted class label is obtained by selecting the index corresponding to the maximum value in the output vector:
\begin{equation}
    \hat{y}_i =
    \begin{cases}
    1, & \text{if } y_i \geq 0.5 \\
    0, & \text{otherwise}.
    \end{cases}
    \label{eq:binclass_prediction}
\end{equation}
In the one-vs-all binary classification framework, each digit class (0-9) is classified against the remaining digits individually, resulting in ten separate binary classifiers. 
This approach allows us to analyze class-wise performance and model generalization for individual digits. 
The use of ridge regression to determine output weights using closed-form matrix computations is a very fast training approach and avoids problems with local minima associated with non-invertible optimization problems encountered for other ANN concepts.   
To optimize the performance of both, the multiclass and the binary classifier, the regularization parameter \(\alpha\) in the ridge regression step is set to \(1.0\). 
This parameter controls the trade-off between minimizing the training error and the magnitude of the weights, thus directly influencing the generalization ability of the model. 

In order to compare the classification performance with a more sophisticated and accurate model, we train a deep neural network (DNN). 
DNNs consist of multiple layers of interconnected units known as artificial neurons, which process information and make predictions based on learned patterns in data. 
Each neuron receives multiple input signals \((x_1,x_2,x_3,...,x_n)\), multiplies them with trainable weights \((w_1,w_2,w_3,...,w_n)\), adds a bias term, and passes the result through a nonlinear activation function ReLU. 
This operation produces an output signal \((\hat{y})\), which contributes to the prediction made by the network. This process is illustrated in Figs. \ref{fig:ML-models}(c) and \ref{fig:ML-models}(d). 
Figure \ref{fig:ML-models}(c) shows a fully connected feedforward network with trainable weights, and \ref{fig:ML-models}(d) depicts the forward computation and gradient-based learning using Adam optimizer.
The DNN model helps us to explore the benefits of nonlinear feature extraction and multi-layer iterative learning. 
Our feedforward DNN is built using TensorFlow and Keras, two well known open source libraries for machine learning. 
The network architecture begins with an input layer that corresponds to the dimension of input data, which is one-dimensional (1D) matrix of 286 values, followed by three hidden layers with decreasing neuron counts to extract meaningful features, which is a common practice. 
Each hidden layer incorporates the ReLU activation function\cite{Bai2022} to introduce non-linearity and enhance learning capacity. 
The output layer of the model is activated by the softmax function, which maps the final feature representation to a set of predefined classes, allowing the network to predict the most likely category for each input instance\cite{Yang2016, Krizhevsky2017}. 
For efficient model training, we implement the Adam optimizer\cite{Kingma2014}, which combines momentum and adaptive learning rates for faster convergence. 
Adam uses error back propagation to compute gradients and updates network weights by minimizing the sparse categorical cross-entropy loss. 
This makes it effective for tasks involving multi-class classification with integer-labeled data. 
Classification performance is measured in terms of accuracy and loss.

\section{Results and Discussion}
In this section we evaluate the classification performance of our machine learning models on the experimental data. 
We test the performance of our SPIC system against the complete Modified National Institute of Standards and Technology (MNIST) database. 
This classic benchmarking image classification dataset contains grayscale images of handwritten digits ranging from 0 to 9. 
The dataset contains a total of 70,000 images, with 60,000 allocated for training and 10,000 for testing. 
Each image contains one digit with a size of \(28 \times 28\) pixels that is first binarized and then scaled to map onto the full surface of the DMD. 

\begin{figure}[htbp]
\centering\includegraphics[width=13cm]{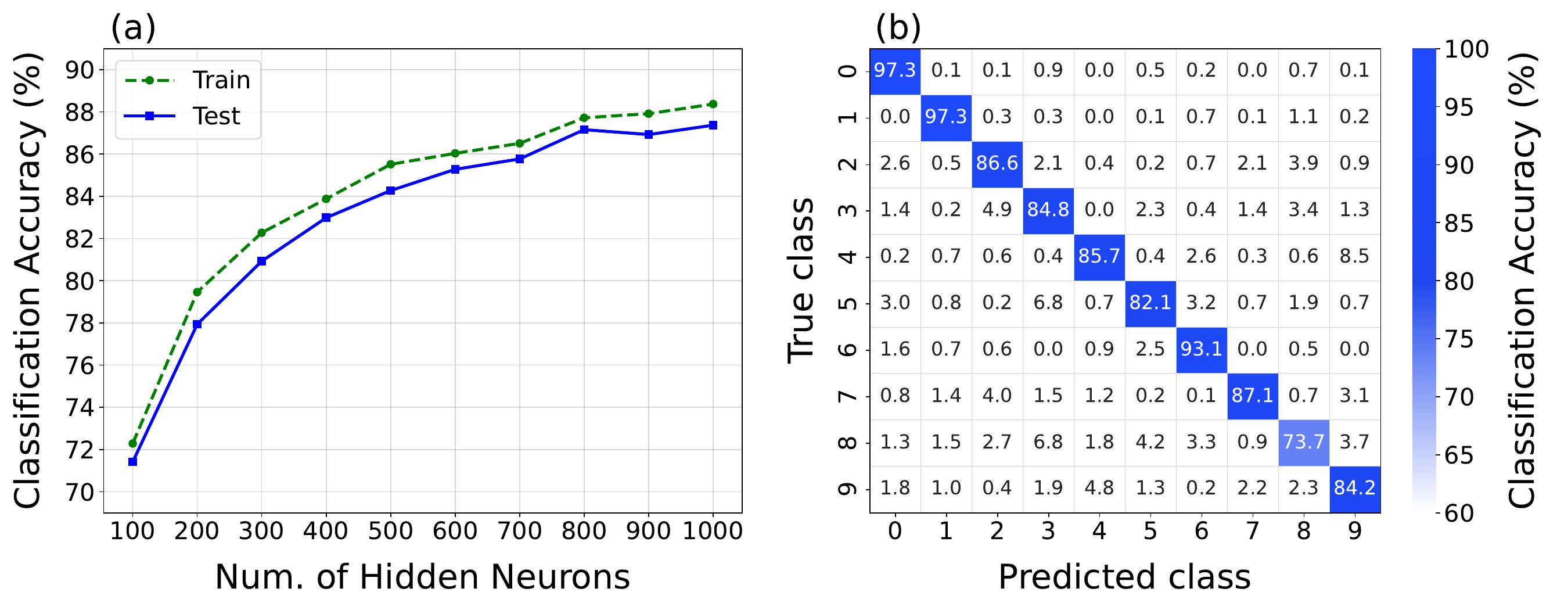}
\caption{(a) Classification performance of the multiclass ELM model versus hidden neurons at \(\alpha = 1\). 
(b) Normalized confusion matrix (\%) for the multiclass ELM model using \(1000\) hidden neurons at \(\alpha = 1\).}
\label{fig:ELM-multiClassification}
\end{figure}

We first present the multiclass and one-vs-all binary classification results obtained from the ELM model, which also serves as a baseline for later comparison with the DNN model. 
Figure \ref{fig:ELM-multiClassification} summarizes the performance of the ELM based classifier averaging the ten digit classes.
Figure \ref{fig:ELM-multiClassification}(a) shows the training and testing accuracies as a function of the number of neurons in the hidden layer. 
The classification accuracy steadily increases with the number of neurons in the hidden layer. 
However, as the curve shows a trend to saturation for high numbers of neurons, we stopped our analysis at 1000 neurons. 
The difference between training and test accuracy remains below 1\%. 
This indicates that our ELM model is able to generalize its classification to data that has not seen before with minimal overfitting. 
Figure \ref{fig:ELM-multiClassification}(b) depicts the normalized confusion matrix for the 10 MNIST classes in the case of 1000 neurons in the hidden layer. 
This confusion matrix is computed on the test data set (10,000 images) to evaluate the multiclass classification. 

\begin{figure}[htbp]
\centering\includegraphics[width=13cm]{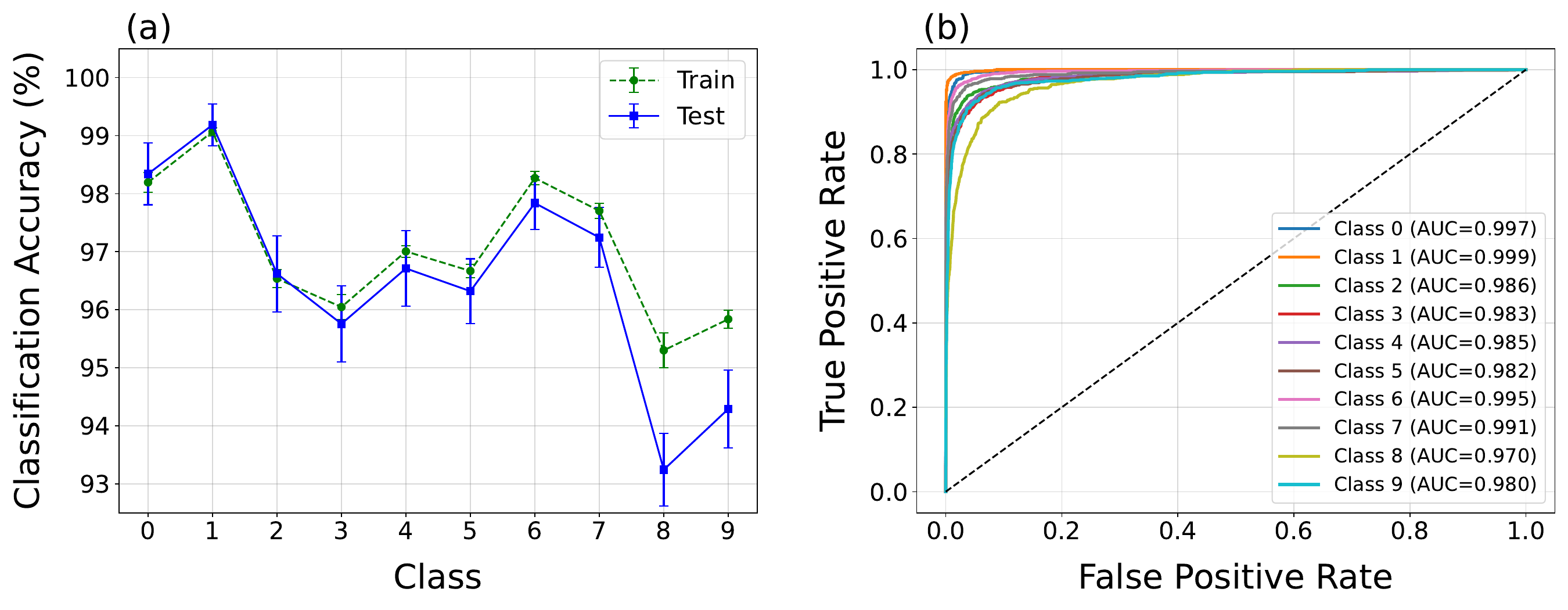} 
\caption{ 
(a) Classification performance of the one-vs-all ELM binary classifier at \(\alpha = 1\) with \(1000\) hidden neurons. (b) Receiver operating characteristics (ROC) curves for each class in binary ELM classifier. 
The different values for the area under the curve (AUC) are all close to 1, indicating strong separability for all classes.}
\label{fig:ELM-ROC-confusionMatrix}
\end{figure}

While the multiclass ELM shows stable and balanced performance, it does not provide insight into how well individual digit classes are separated. 
To analyze class-wise performance in more detail, we use one-vs-all binary ELM classifiers. 
Figure \ref{fig:ELM-ROC-confusionMatrix} shows the performance of binary classifiers at \(\alpha = 1\) using 1000 hidden neurons. 
Panel (a) depicts the training and test accuracies for each digit class. 
The gap existing between training and testing accuracies, particularly for classes 8 and 9, indicates the limits of generalization performance towards new data in our model. 

When we train our ELM model for binary classification, it does not simply return a yes or no for every image but rather assigns a confidence score to every prediction. 
For instance, it may return, "80\% certain this is a 3", or "60\% certain this is a 9". 
In order to make an actual decision, we define a threshold (e.g. 50\%) and if the result is above that threshold, we declare it a match. 
It is clear that different thresholds can produce varying outcomes. 
To understand the performance of a binary classifier, it is important to capture the classifier's discriminative ability across thresholds. 
To provide a more comprehensive evaluation, we plot the receiver operating characteristics (ROC) curves and calculate the corresponding areas under the curve (AUC) for our classifier in Fig. \ref{fig:ELM-ROC-confusionMatrix}(b). 
The use of the ROC-AUC curve allows us to visualize how well the classifier distinguishes between the positive class (target class) and the negative class (all other classes) at different confidence levels. 
The ROC curve plots the true positive rate (the number of times the positive class is correctly identified) shown in equation (\ref{eq:TPR}) against the false positive rate (the number of times other classes are incorrectly classified as positive class) given in equation (\ref{eq:FPR}) across a range of classification thresholds. Those rates are defined as: 

\begin{equation}
\text{TPR (True Positive Rate)} = \frac{\text{TP}}{\text{TP} + \text{FN}}
\label{eq:TPR}
\end{equation}

\begin{equation}
\text{FPR (False Positive Rate)} = \frac{\text{FP}}{\text{FP} + \text{TN}}.
\label{eq:FPR}
\end{equation}

\noindent Here, TP (true positives) are correctly identified samples of the target class, FP (false positives) are misclassified non-target samples, FN (false negatives) are missed target samples, and TN (true negatives) are correctly identified non-targets. 
AUC values close to 1 indicates excellent separability, whereas a slope of 0.5 corresponds to random guessing. 
Our result shows consistently high AUCs across most classes, confirming strong discriminative power of our binary classifier. 
Slightly lower AUCs for some classes (e.g. class 8), align well with the overfitting trend as explained earlier. 


We next evaluate the performance of the DNN classifier. 
There exists a trade-off between projection bandwidth (i.e. inverse of the time required to send the complete set of patterns to reconstruct an image) and the amount of spatial information recovered. 
Using fewer patterns speeds up the acquisition process but leads to a lower accuracy in the classification (due to a lower resolution in the final reconstructed image). 
Starting from the complete set, we study how by optimizing the selection of pattern subsets, we can maintain good classification accuracy while increasing the effective bandwidth. 

\begin{figure}[htbp]
\centering\includegraphics[width=0.6\textwidth]{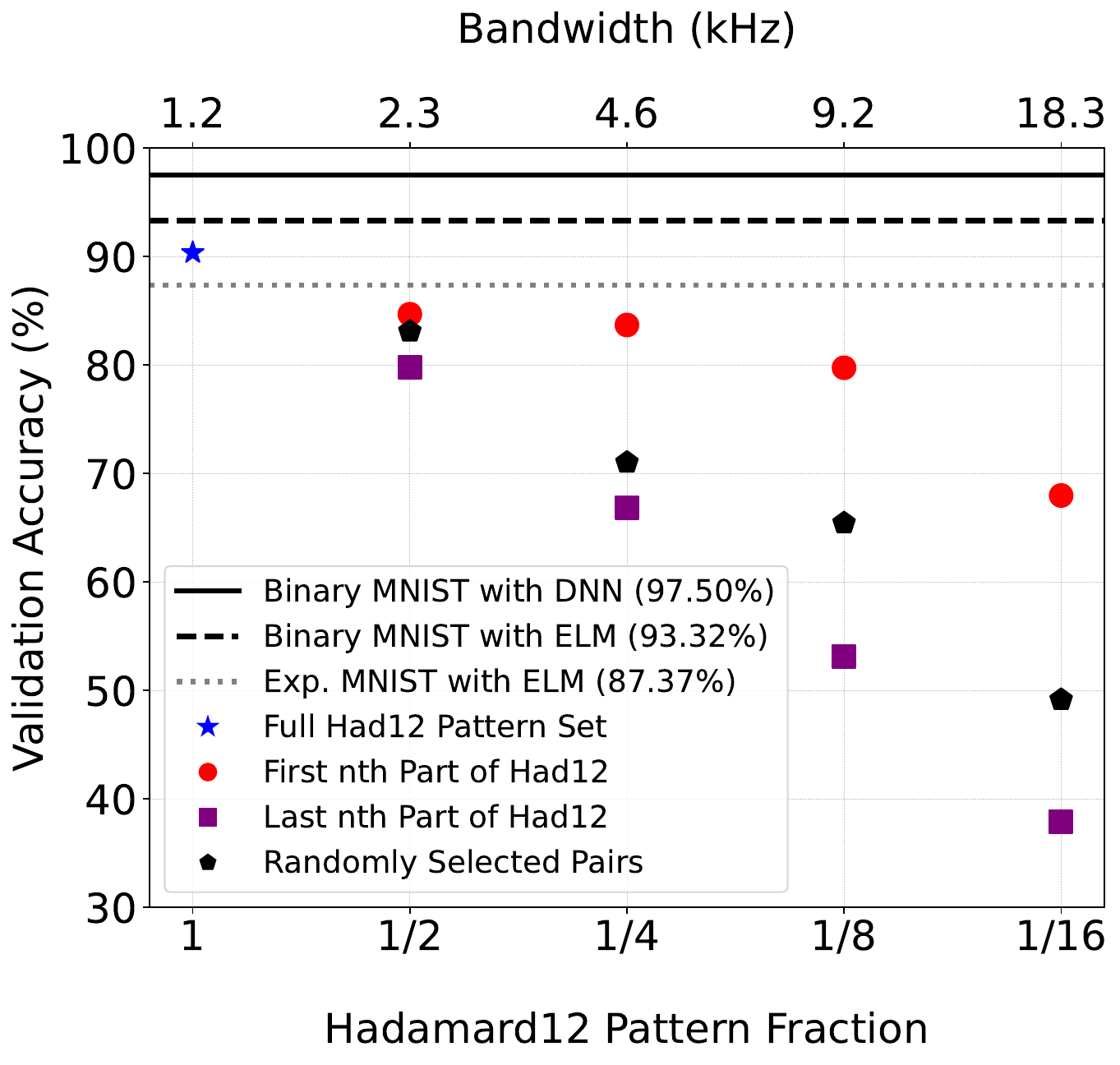}
\caption{Validation accuracy for different subsets of Had12 after model training for 300 epochs. 
The bandwidths at which the images are classified are depicted on the top axis.}
\label{fig:had12compareperform}
\end{figure}

Figure \ref{fig:had12compareperform} compares the performance of our SPIC for different subsets of the Had12 pattern set (144 patterns in total). 
The bottom x-axis represents the fraction of pattern set used in each case, with the corresponding inference bandwidth in kilohertz (kHz) indicated on the top x-axis. 
Horizontal reference lines indicate baseline performances: Binarized MNIST numerically classified with a DNN (black solid line, 97.50\%), binarized MNIST numerically classified with ELM (black dashed line, 93.32\%), and experimentally encoded MNIST with Had12 pattern set classified with ELM (gray dotted line, 87.37\%).
Using the full Had12 pattern set on the DNN (blue star) achieves classification accuracy above 90\%, outperforming the experimental ELM baseline, and closely matching the performance obtained with numerically simulated binarized MNIST dataset. 
We observe in Fig. \ref{fig:had12compareperform} that the validation accuracy decreases as soon as we reduce the number of patterns used from the original Had12 set. 
Moreover, the classification performance varies based on the pattern selection strategy. 
The first n$_{\mathrm{th}}$ part of Had12 (red circles) consistently maintains the highest accuracy. 
There is an ordinal hierarchy in the generation of the Hadamard patters (cf. \ref{appendix:A}), with the number of changes between $\pm1$ (inversions) increasing with the ordinal pattern number. 
Our result clearly shows that the Had12 patterns with lower number of inversions contain more useful information for accurate classification. 
In clear contrast, last n$_{\mathrm{th}}$ part of Had12 (purple squares), corresponding to high number of inversions, achieve lower overall accuracies and performance decays faster with the fraction. 
Consistently, the randomly selected pairs (black pentagons) show intermediate performance. 
This finding suggests a clear strategy to reduce the number of Had12 patterns, which increases the effective bandwidth, while keeping relatively high image classification accuracies. 
Using the full Had12 pattern set, 10,000 optically processed MNIST digits can be classified by the DNN in less than 0.73 seconds (inference time of 73 \(\upmu\)s/digit), while this decreases to 0.31 seconds for 10,000 MNIST digits in the ELM model (inference time of 31 \(\mu\)s/digit). 
The ELM's inference time is 2X faster than the DNN, but the accuracy of the latter is higher. 

\begin{figure}[htbp]
\centering\includegraphics[width=\textwidth]{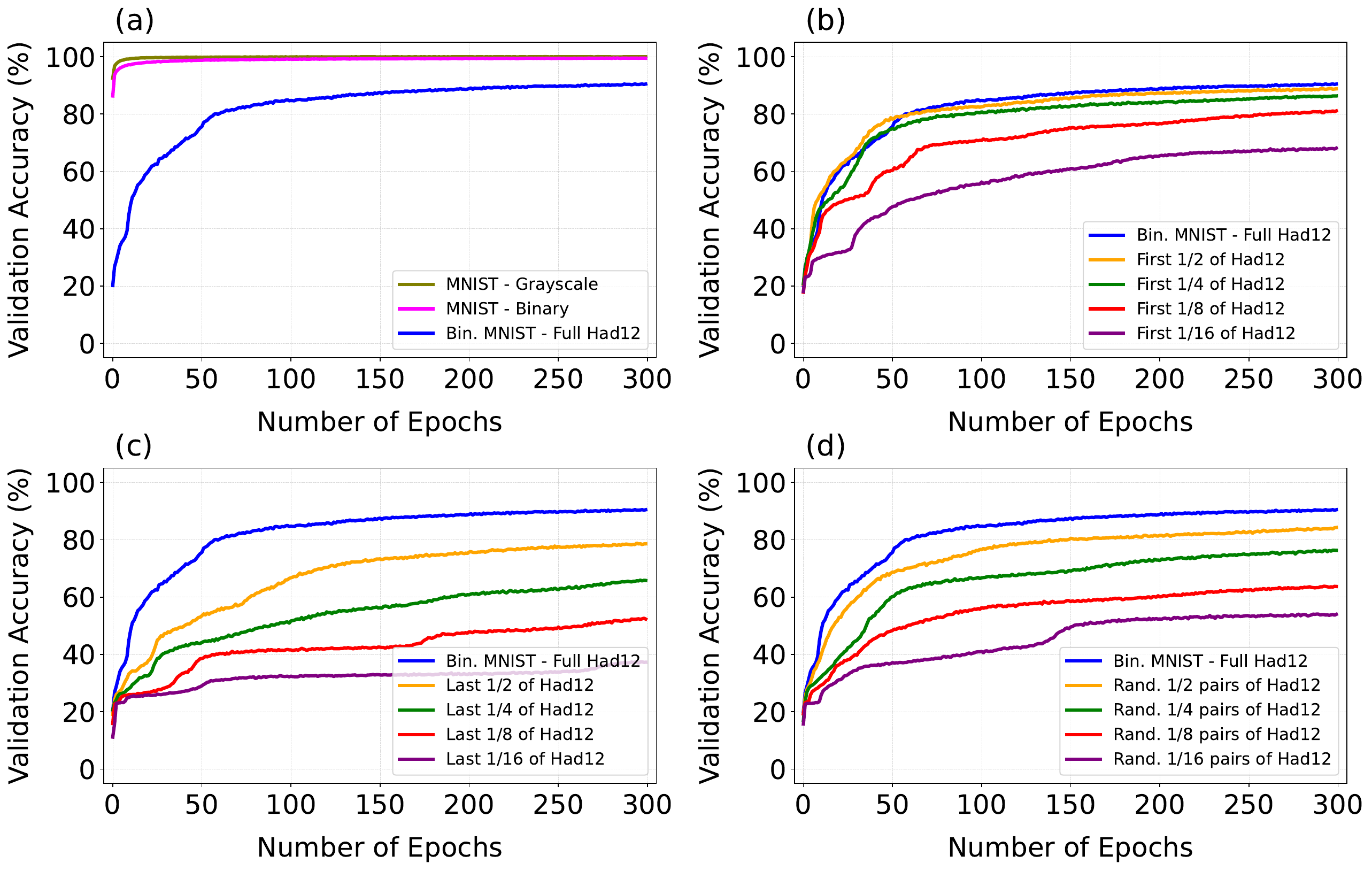}
\caption{Validation accuracy learning curves over 300 epochs for a DNN on MNIST variants and Had12 subsets. 
(a) Numerical simulation of the original (grayscale) MNIST and binarized MNIST, and experimental MNIST encoded with full Had12 set. 
(b) Binary MNIST encoded with the full Had12 and with the first $\tfrac{1}{2}$, $\tfrac{1}{4}$, $\tfrac{1}{8}$, and $\tfrac{1}{16}$ patterns of the Had12 set. 
(c) Binary MNIST encoded with the full Had12 and with the last $\tfrac{1}{2}$, $\tfrac{1}{4}$, $\tfrac{1}{8}$, and $\tfrac{1}{16}$ patterns of the Had12 set. 
(d) Binary MNIST encoded with the full Had12 set and with randomly selected $\tfrac{1}{2}$, $\tfrac{1}{4}$, $\tfrac{1}{8}$, and $\tfrac{1}{16}$ pairs of the Had12 set.}
\label{fig:trainingCurves}
\end{figure}

In order to better understand the impact of Had12 subsets in the DNN's validation accuracy, we study in detail the learning curves of the DNN for the different cases. 
We plot in Fig. \ref{fig:trainingCurves} the validation accuracy learning curves over 300 epochs. 
Figure \ref{fig:trainingCurves}(a) depicts the learning curves for the original grayscale (olive) and binarized (magenta) MNIST datasets, which smoothly converge to high accuracies in a few learning steps as expected for numerically simulated data. 
In contrast, the experimental binarized MNIST encoded with the complete Had12 presents a slower rise saturating at an accuracy $\gtrsim90\%$. 
Figures \ref{fig:trainingCurves}(b), \ref{fig:trainingCurves}(c) and \ref{fig:trainingCurves}(d) highlight the effect of using different fractions of Had12 patterns: As discussed above, when the number of patterns decreases ($\tfrac{1}{2}$, $\tfrac{1}{4}$, $\tfrac{1}{8}$, and $\tfrac{1}{16}$), the final accuracy drops significantly.
Notably, those training curves present a higher impact of vanishing gradients, i.e. phases of stagnated growth of the validation accuracy, the lower the fraction of patterns used in the classification. 
During the periods of vanishing gradient the network worsens its generalization capacity. 
As training progresses, the optimizer eventually escapes these shallow minima and accuracy grows again. 
This behavior is common in deep learning, particularly when training with compressed inputs such as sub-sampled Hadamard patterns\cite{Fan2021, Petzka2021}. 
Figure \ref{fig:trainingCurves}(c) corresponds to the last patterns of the Had12 sequence and shows the longest periods of vanishing gradients, which is consistent with the resulting lower accuracies shown in Fig. \ref{fig:had12compareperform}. 

At this point, the balance between image classification bandwidth and validation accuracy has been studied in detail. 
There is still however the open question of what is the role of the image quality, i.e. its signal-to-noise ratio in the learning capacity of our DNN. 
Next, we explore how noise affects the learning and validation accuracy of our DNN following the general framework introduced in \cite{Semenova2024}, specifically focusing on injecting additive uncorrelated Gaussian noise. 
Here, we inject additive uncorrelated noise directly into the input layer of our feedforward DNN, resembling a uniform loss in image quality. 
Each input neuron receives an independent noise value, ensuring that noise is added separately to every input feature. 
Let $X_i$ denote the $i$-th input feature to the network. 
Therefore, to simulate noisy inputs, we modify the input layer as follows:

\begin{equation}
\tilde{X}_i = X_i + \sqrt{2D_A^U} \cdot \xi_A^U(t, i).
\label{eq:input-noise}
\end{equation}

\noindent Here, $D_A^U$ denotes the noise intensity, the subscript $A$ refers to the additive noise, and superscript $U$ indicates that the noise is uncorrelated. 
$\xi_A^U(t, i) \sim \mathcal{N}(0,1)$ is a zero-mean, unit-variance Gaussian noise term that is uncorrelated across input dimensions. 
This models a realistic scenario in which each input feature experiences independent noise at each time step. 

\begin{figure}[htbp]
\centering\includegraphics[width=\textwidth]{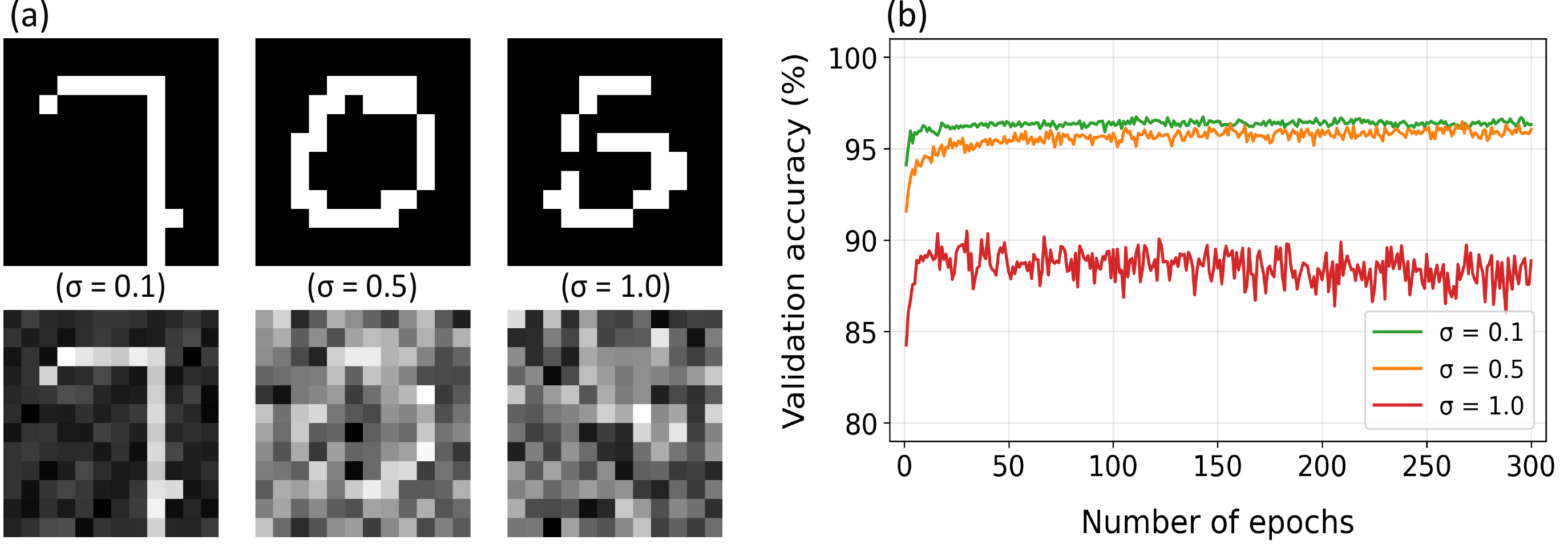}
\caption{(a) Examples of MNIST digits injected with additive uncorrelated Gaussian noise at various noise intensities \(\sigma = 0.1, 0.5, 1.0\). (b) Numerical validation accuracy of a feedforward DNN on binary MNIST dataset. Additive uncorrelated Gaussian noise is injected in the input layer, where \(\sigma = \sqrt{2D}\) shows the noise intensity.}
\label{fig:white-noise}
\end{figure}

To analyze the impact of Gaussian noise, we have chosen three different levels of noise intensity denoted by \(\sigma\). 
Figure \ref{fig:white-noise}(a) top row shows some examples of binarized MNIST digits, and bottom row shows the same digits after adding Gaussian noise at different noise intensity levels. 
When the noise level is low \(\sigma = 0.1\), the digit structure remains mostly clear and readable. 
With the increase in noise level \(\sigma = 0.5,\) and \(1.0\), the digits become increasingly blurred and, in the case of the highest noise level, barely identifiable by a human observer. 
Figure \ref{fig:white-noise}(b) shows the impact of this additive uncorrelated Gaussian noise on overall classification performance over 300 epochs.
For noise intensities \(\sigma = 0.1\) and \(0.5\), the model performs reasonably well and converges to accuracies above 95\%.
However, for \(\sigma = 1.0\), the validation accuracy drops significantly and fluctuates more, shows that high noise at the input layer makes it harder for the model to distinguish between classes reliably. 
We observe that the addition of Gaussian noise to the input data degrades the validation accuracy as originally expected. 
However the learning curves quickly converge to their asymptotic values and this is in stark contrast to the type of learning curve convergence observed in the compressed sensing scenario, where the learning algorithm gets stuck in sub-optimal local minima for longer the less information is used to classify the images. 
We therefore conclude that the main effect originating the degradation of the performance in our experiment is the lack of structured information after depriving the learning algorithm of specific parts of the Had12.  

\section{Conclusions}
We experimentally demonstrated SPI based MNIST image classification with accuracies above $90\%$ at 1.2kfps using a microLED digital light projector and \(12 \times 12\) Hadamard patterns.
We have studied the classification accuracy versus the fraction of Had12 used in the projection and we have demonstrated that the optimal approach to reduce the number of Had12 used is by categorizing the patterns based on their spatial frequency content. 
Following this approach we drastically reduced the number of projected patterns to the first 1/4 of the original set, while keeping the classification accuracy to $\simeq 78\%$. 
Further, we numerically studied how white Gaussian noise affects the classification performance. 
We were able to show that the decrease in performance for our DNN learning algorithm is not related to a change in equivalent signal-to-noise ratio when using less symbols to reconstruct the image but is bound to the loss of spatial information under compressed sensing. 
Finally, we demonstrated the possibility to use a simple nonlinear ELM model together with our SPI experiment as binary classifier for the MNIST dataset. 
This result provides a foundation for future use of SPIC in anomaly detection problems in machine vision systems. 


\appendix
\renewcommand{\thesection}{Appendix~\Alph{section}}
\section{The Hadamard patterns sequence}
\label{appendix:A}

The Hadamard set forms a complete orthogonal basis of binary patterns derived from the original Hadamard matrices, which are square matrices consisting of elements valued +1 and -1 \cite{Assmus1992}. 
The Kronecker product of two Hadamard matrices \( H_F \) and \( H_B \) is also a Hadamard matrix: \( H_{F \cdot B} = H_F \otimes H_B \). 
A row (or column) of \(H_N\) can be characterized by its \textit{sequence}, which refers to the number of transitions from +1 to -1, or vice-versa in the pattern. 
This concept is analogous to the frequency of a wave in Fourier analysis\cite{Herman2015}. 
A complete Hadamard set comprises as many patterns as there are pixels in the reconstructed images (sampling rate = 1). 
In this work, we employ a \(12 \times 12\) Hadamard pattern set (Had12). 
The following algorithm is used to generate the sequence of Hadamard patterns subsequently used in the SPI experiment. 

\begin{algorithm}[H]
\caption{Generation of Hadamard Pattern Set}
\label{alg:had_pats}
\begin{algorithmic}[1]

\Require Integer dimensions $n,m$ (typically $n=m\in\{12\}$)
\Ensure Pattern sequence $\mathcal{P}$ and complementary inverse sequence $\mathcal{P}^{-}$

\If{$m$ is not provided}
    \State $m \gets n$
\EndIf

\State Generate Hadamard matrices $H_N \gets \texttt{hadamard}(n)$ and $H_M \gets \texttt{hadamard}(m)$
\State Initialize empty buffer $\texttt{buf} \gets \{\}$

\For{$i = 1$ to $n$}
    \For{$j = 1$ to $m$}
        \State Construct 2D Hadamard-derived pattern
        \[
            P_{i,j} = H_N(:,i)\,H_M(j,:)
        \]
        \State Append $P_{i,j}$ to $\texttt{buf}$
    \EndFor
\EndFor

\State Remove DC pattern: discard first element of $\texttt{buf}$

\State Initialize $\mathcal{P} \gets \{\}$ and $\mathcal{P}^{-} \gets \{\}$

\For{each pattern $P_k$ in $\texttt{buf}$}
    \State Append $P_k$ to $\mathcal{P}$
    \State Compute inverse $P^{-}_k \gets -P_k$
    \State Append $P^{-}_k$ to $\mathcal{P}^{-}$
\EndFor

\end{algorithmic}
\end{algorithm}

\begin{figure}[htbp]
\centering\includegraphics[width=13cm]{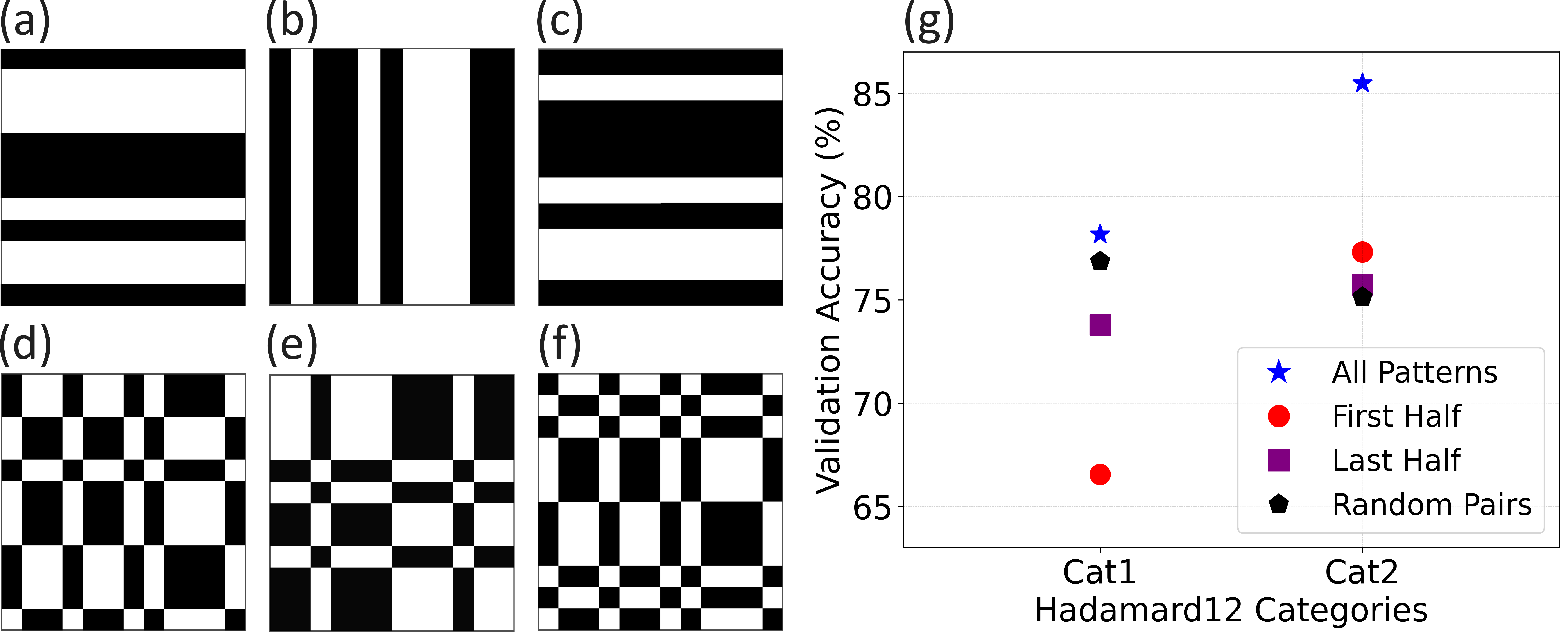}
\caption{Examples of Hadamard patterns with binary intensity values encoded in white (+1) and black (-1). Panels (a), (b) and (c) correspond to patterns in category 1 showing one-dimensional spatial distribution only (cf. explanation in the main text). 
Panels (d), (e) and (f) illustrate patterns in category 2, i.e. showing two-dimensional spatial distribution. (g) Validation accuracy of the experimental binarized MNIST using only Hadamard12 categories over 300 training epochs.}
\label{fig:cat1&2-patterns-class}
\end{figure}

Figure \ref{fig:had12compareperform} depicts a large gap between the classification accuracies for the first and last fractions of Had12 patterns. 
This motivates a more detailed study based on the internal structures of Had12 patterns. 
Based on their internal structure ---the number of transition between +1 and -1 across spatial dimension---, we can split the Hadamard patterns in two distinct categories. 
For a \(12 \times 12\) Hadamard pattern base as shown in Fig. \ref{fig:cat1&2-patterns-class}:
\begin{itemize}
    \item \textbf{Category 1 (Cat1):}  The first 44 patterns, these exhibit sign changes along only one spatial axis (either horizontal or vertical), and capture basic, low-detailed features.
    \item \textbf{Category 2 (Cat2):} The remaining patterns (indices 45–288), which change in both spatial directions and capture finer, more detailed features.
\end{itemize}


Figure \ref{fig:cat1&2-patterns-class} illustrates the classification performance of two categories of the Had12 pattern set. 
Since, Had12 patterns are not evenly distributed, approximately 80\% of the Had12 patterns belong to Cat2, while only 20\% fall under Cat1. 
This imbalance plays a key role in performance.  
In most sampling strategies, Cat2 achieves slightly higher classification accuracy mainly because it contains a broader and denser set of patterns.
Despite this, Cat1 shows competitive performance even though it represents only a small fraction of the entire pattern set. 
This indicates that low-sequency patterns alone carry useful features, which is often sufficient for capturing the coarse structure required for simple classification tasks.
The plotted results compare experimental validation accuracies for various subsampling strategies across both categories including all patterns, first half, last half, and half random pairs. 

\begin{backmatter}
\bmsection{Funding}
Engineering and Physical Sciences Research Council (EP/M01326X/1, EP/S001751/1, EP/T00097X/1, EP/V004859/1); Royal Academy of Engineering (“Research Chairs and Senior Research Fellowships”); The Volkswagen Foundation. 
\bmsection{Data Availability Statement}
The data for this manuscript is publicly available in the following link \cite{OpenDataSPIC}.
\end{backmatter}

\bibliography{sample}
\end{document}